# A Noise and Edge extraction-based dual-branch method for Shallowfake and Deepfake Localization

Deepak Dagar[1], Dinesh Kumar Vishwakarma[2,*]

Biometric Research Lab, Department of Information Technology(IT), Delhi Technological University(DTU), Bawana Road, Delhi-110042, India

deepakdagargate@gmail.com[1], dvishwakarma@gmail.com[2,*]

**Abstract**

The trustworthiness of multimedia is being increasingly evaluated by advanced Image Manipulation Localization (IML) techniques, resulting in the emergence of the IML field. An effective manipulation model necessitates the extraction of non-semantic differential features between manipulated and legitimate sections to utilize artifacts. This requires direct comparisons between the two regions.. Current models employ either feature approaches based on handcrafted features, convolutional neural networks (CNNs), or a hybrid approach that combines both. Handcrafted feature approaches presuppose tampering in advance, hence restricting their effectiveness in handling various tampering procedures, but CNNs capture semantic information, which is insufficient for addressing manipulation artifacts. In order to address these constraints, we have developed a dual-branch model that integrates manually designed feature noise with conventional CNN features. This model employs a dual-branch strategy, where one branch integrates noise characteristics and the other branch integrates RGB features using the hierarchical ConvNext Module. In addition, the model utilizes edge supervision loss to acquire boundary manipulation information, resulting in accurate localization at the edges. Furthermore, this architecture utilizes a feature augmentation module to optimize and refine the presentation of attributes. The shallowfakes dataset (CASIA, COVERAGE, COLUMBIA, NIST16) and deepfake dataset Faceforensics++ (FF++) underwent thorough testing to demonstrate their outstanding ability to extract features and their superior performance compared to other baseline models. The AUC score achieved an astounding 99%. The model is superior in comparison and easily outperforms the existing state-of-the-art (SoTA) models.

## 1 Introduction

Advancements in computer graphics and deep learning have provided individuals with enhanced capabilities to generate deceptive visuals. Thanks to the progress of advanced editing AI tools, it is now feasible to effortlessly modify multimedia data to create extremely lifelike content. Manipulated media can cause significant harm, including concrete damage, psychological distress, and physical suffering. A deceitful individual may employ the technology to illicitly obtain something valuable with the intention of causing harm [1]. Image-editing tools or graphics-based procedures that follow traditional approaches are commonly referred to as "**shallow fakes**" [2]**.** Shallowfakes have been somewhat altered, but not to the extent of or sophistication of "**deepfakes**" Deepfakes refers to approaches based on deep learning. Deepfake has garnered significant interest due to its ability to generate exceptionally realistic and convincing content with remarkable ease. As advanced technologies progress, there is a need for more sophisticated image modification localization approaches to address existing modified images and mitigate security risks. The objective of the localization challenge is to detect and delineate the changed areas in an image with pixel-level precision. Shallow fakes involve three distinct forms of image manipulation. Shallowfakes can be divided into three categories [2](Figure I):

1) Splicing: Copying a portion of an image and transferring it into another.
2) Copy-move: Replicating a specific portion of an image.
3) Removal: removing portions of images or an object is removed from the image.

Furthermore, Deepfake can be categorized into three distinct categories [1]:

1) Face Swap: The act of transferring a person's face from one image or frame to another.
2) Face Reenactment: It involves transferring facial expressions or movements from a source to a target.
3) Entire Image Synthesis: The complete image is generated using advanced AI technologies.

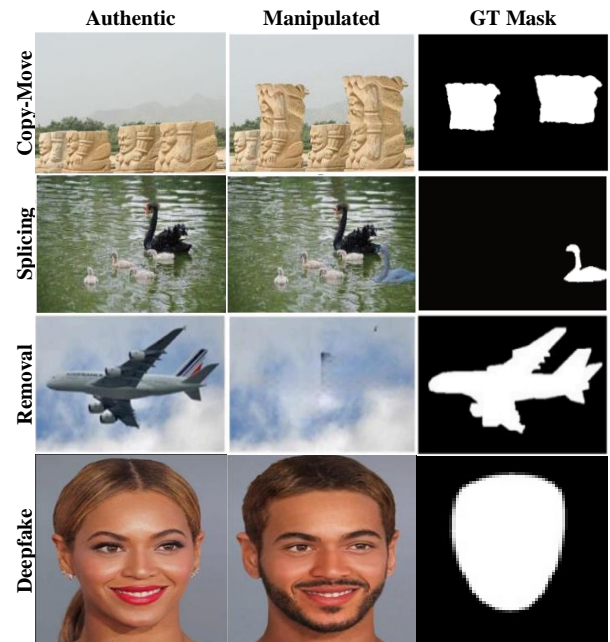

Figure I: Illustration of various types of image manipulation

Handcrafted feature-based approaches often operate under the assumption that a specific manipulation artifact is present. Their objective is to expose the counterfeits by examining regional inconsistencies in the color filter array (CFA) [3] [4], illumination [5], photo response non-uniformity noise(PRNU) [6], compression artifacts in JPEG images [7], texture units [8]. Nevertheless, these solutions presuppose prior knowledge of the tampering, hence restricting their effectiveness in many situations. Deep learning methods, specifically CNNs have a tendency to acquire contextual (semantic) information and employ advanced strategies for detecting image counterfeiting [9] [10] [11] resort to them [12]. Convolutional neural networks (CNNs) are based on local information and focus on the relationship between surrounding regions. However, they may not capture the global interactions between different parts of the image. Artifacts are discrepancies in imperceptible low-level characteristics, such as noise or high-frequency, that are not discernible to the unaided eye but are obvious when examined closely [13]. Therefore, according to the previous study, the crucial aspect of the manipulation task is to identify minor discrepancies and visible evidence that are unrelated to meaning to detect artifacts. Handcrafted features are limited since they presuppose the sorts of tampering before and may not detect other types of tampering. On the other hand, Convolutional Neural Networks (CNNs) learn semantic features and analyze links between nearby artifacts, but this alone is insufficient for learning manipulation artifacts.

To overcome these constraints, we have created a dual-branch model that leverages handcrafted features, such as RGB information on one branch and noise inconsistencies on another using the aid of edge supervision and the hierarchical structure of ConvNext. The contribution of the paper is summarized below:

➢ Developed an innovative dual-branch structure that utilizes RGB data in one branch and noise characteristics in another using the hierarchical structure of ConvNext, with the features being improved via a feature enhancement module. The model also utilizes edge supervision to improve the localization of manipulation at the edges.
➢ Experimentation is performed on the shallowfakes dataset, which comprises NIST16, Coverage, Columbia and CASIA datasets, along with the deepfake dataset Faceforensics++. The results indicate that the FF++ model outperforms other models in terms of its discriminative capacity.
➢ Qualitative analysis is conducted to visually examine the localization result of the model and compare them with the visualization results of the state-of-the-art (SOTA) models.
➢ Ablation studies are performed to investigate the significance of various components within the overall model.

## 2 Related Work

Image Localization Methods(IML) approaches can be classified into two primary categories according to feature representation: handmade features and methods based on deep learning.
. This section will explore these two categories.

### 2.1 Handcrafted features

Handcrafted feature-based solutions traditionally model authentic images to disclose the statistical relationship between pixels and capture the statistical variation caused by image manipulation procedures. For example, the spatial rich model (SRM) [14], the technique, commonly employed in image steganalysis, was extended in [15] to obtain residual-based features for the multidimensional Gaussian model and SVM classifier to detect and locate image fraud. Nevertheless, modifying an image will invariably lead to modifications in its visual elements, which might be utilized by local image describers to identify counterfeit versions. To identify the alterations caused by image splicing, [5] combined the statistical data obtained from various local descriptors that analyze texture, illumination, shape, and color characteristics. Image alteration impacts the non-semantic characteristics and leads to low-level features such as noise inconsistencies. Researchers have further explored these features to obtain more effective discriminative features. Zhang et al. [16] utilize an enhanced version of the constrained convolutional model to extract noise features that act as more nuanced indicators of manipulation. These features are then inputted into a dual-branch architecture for additional feature learning. Nevertheless, these noise characteristics exhibit inconsistency and resilience when subjected to JPEG compression and Gaussian blur.

Li et al. [17] use a Markov-based method in the Quaternion Discrete Cosine Transform (QDCT) domain. This method expands on the Markov transition probability characteristics found in QDCT frequency domains to reveal the interconnection between adjacent pixels. Nevertheless, the technique is restricted to color photographs, and its effectiveness is uncertain due to different instances of compression blurring. Handcrafted features assume that hidden manipulation characteristics can be observed in certain manipulation artifacts, and a technique specifically intended to capture these artifacts will yield superior outcomes. Nevertheless, these methods have limited ability to be applied to a wide range of situations, and they lack the necessary strength to handle many types of manipulation operations.

### 2.2 Deep learning-based method

Deep learning algorithms can autonomously train and optimize feature representations for purpose of forgery forensics. This is different from conventional techniques, which rely on a tedious feature engineering process to manually construct features. Wu et al. [11] suggest a method called Mantra-Net is designed for localizing and detecting generalized image forgeries

(IFLD). The system is capable of detecting specific irregularities in the image that can indicate the presence of manipulated pixels. It offers a comprehensive approach to detecting different types of forgeries. They are eliminating the need for pre- and/or post-processing. Nevertheless, the model is limited in its ability to handle multi-forgeries and images with strongly correlated noises. Cozzolino et al. [18] introduced an improved initialization method and employed a Siamese network to perform splicing localization and detection. The utilization of Siamese networks in Noiseprint allowed for the detection and identification of camera model artifacts through the analysis of noise residual, aiding in the localization of fraudulent activity [18].

An important obstacle in the field of image manipulation detection is the identification of distinct and adaptable characteristics that can distinguish between real images and manipulated ones, without generating false alarms on genuine images, while yet being able to detect manipulations in new and unfamiliar data. Chen et al. [19] in their study, the authors tackle both aspects by employing multi-scale supervision and multi-view feature learning. The former method aims to obtain characteristics that are not dependent on specific meanings and, thus, may be applied more broadly. This is achieved by analyzing the boundary artifacts and noise distribution surrounding tampered regions. The latter allows us to acquire knowledge from genuine images that are excessively intricate for the current semantic segmentation network-based methods to manage. The model exhibits superior performance in cross-dataset scenarios and demonstrates resilience to diverse post-processing techniques. A multitask fully convolutional network (MFCN) was built to localize forgery. The training data consisted of ground truth information about the fabricated regions and boundaries [9]. In a recent study [10] researchers constructed a hybrid LSTM and encoder-decoder network to localize pixel-wise forgeries. This technique uses spatial attributes and resampling to accurately capture irregular shifts between counterfeit and genuine patches. CNN approaches excel in comprehending visual context and acquiring semantic characteristics, resulting in superior performance in image classification tasks. Image manipulation is independent of semantic meaning; hence these strategies are generally less successful against image manipulation. CNNs possess inductive and location biases, which prevent them from capturing the global correlation of features.

Objectformer [20] and Transforensics [21] are approaches that utilize transformers and have been proposed in the past. The general architecture and design philosophy of these two models do not mesh well with ViT. Unlike ViT, which directly embeds patched images for encoding, both systems in question utilize many CNN layers to extract feature maps before employing Transformers for further encoding. However, this approach overlooks crucial initial low-level information. In order to address the shortcomings of the current models, we have developed a dual-branch model that incorporates handcrafted features, noise, and features derived from convolutional neural networks (CNNs). The dual-branch architecture comprises one branch that captures RGB information and another branch that extracts noise information using the Bayar convolution and SRM filters. The hierarchical structure of ConvNext passes these features and they are further improved by the feature enhancement module. The model also employs edge supervision, enabling it to concentrate on the border information, where tampering typically occurs. Finally, the features from both branches are added to the mask's final prediction.

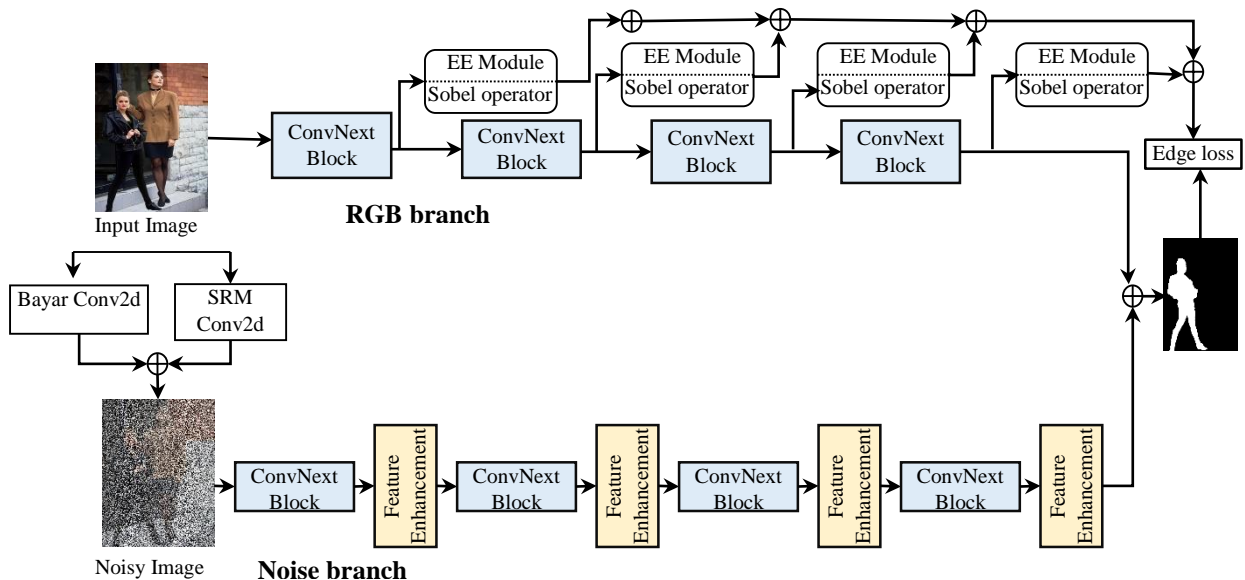

Figure II: Overview of the proposed model consisting of a dual branch consisting of RGB and noise branch followed by ED architecture

## 3 Proposed Model

The proposed model comprises two parallel branches: one branch takes noise/residual characteristics as input, which are determined using the Bayar convolution and SRM convolution filters. These low-level/non-semantic features are further correlated by the ConvNext module and the Feature Enhancement(FE) module further enhances the feature representation by regulating global dependencies along various axes, while the other branch is the contextual branch which uses the spatial characteristics of the image samples to determine manipulation. The Sobel operator and Edge Extraction(EE) module retrieve manipulation edge predictions from the features generated by each ConvNext layer along the contextual branch. Subsequently, the edge extracted features are kept concatenated and used as a supervision for the loss functions(Figure II). The contextual branch is designed to preserve the fine details of the feature and aims to capture more comprehensive correlations among the data. The features obtained from both branches are combined to enhance the precise handling of localized features.

Noise branches and Contextual branches with their components are discussed in great detail in this section.

### 3.1 Noise Inconsistencies

A genuine image has a consistent distribution of noise over the entire image. Different acts of tampering introduce inconsistency and weaken the uniformity in the tampered area. The rationale behind using noise residual as a feature is that the noise characteristics between the source and target images are improbable to be similar when an object is extracted from one image (the source) and inserted into another image (the target). The noise residual in this configuration represents the difference between the true value of a pixel and the estimated value of that pixel, which is calculated by interpolating only the values of adjacent pixels. This estimation acts as the noise model. Various types of noise residual filters or kernels exist, each with its sensitivity to different forms of manipulation. The Bayar and SRM filters are widely used for effectively capturing the residual features of low-level noise.

#### 3.1.1 Bayar Convolution or Constrained CNN

Constrained Convolution Neural Networks can utilize data to learn how image editing operations affect local pixel correlations. Therefore, this method can eliminate material at the image level and dynamically acquire knowledge about the indications of image alteration [22]. Constrained CNN is specifically developed to learn filters that predicts errors, which in turn generate feature maps used as low-level forensic traces. These traces offer improved universality and robustness. Subsequent layers in the neural network would gradually acquire knowledge of the more complex visual features depicted in the low-level traces [23]. In order to constrain the CNN to acquire knowledge of the low-level patterns, specific limitations are imposed on the weights of the CNN's kernel.:

$$\begin{cases} \omega_k^{(l)}(0,0) = -1, \\ \sum_{m,n \neq 0} \omega_k^{(l)}(m,n) = 1, \end{cases} \quad (1)$$

Equation (1) denotes the constraints imposed on the kernel's filter. The superscript indicates the CNN layer. The kth convolutional filter in a layer is represented by the subscript k, and the spatial co-ordinates(0,0) corresponds to the central value of the filter. During training, in the backward pass, the weights are updated using the optimizer. Then, the central value of the kernel is set to zero, and the remaining weights of the kernel are normalized so that their sum is equal to one. Finally, the central value of the filters is updated to minus one.

#### 3.1.2 Steganalysis Features

SRM (Spatial Rich Models) filters are another technique employed to extract features from the noise residuals of an image. Fridrich et al. [14] initially established the concept of Spatial Rich Models (SRM). Its main purpose is to do steganalysis, which involves extracting concealed or hidden characteristics from the noisy residuals of an image using a predetermined set of high-pass filters. Afterwards, the aforementioned features are merged and transmitted to ensemble classifiers. This method is specifically developed to calculate the necessary statistics for extracting specified properties from the noise residuals surrounding the pixel neighborhood in an image. If we view the data embedding process in steganography as a specific form of image tampering, then image forensics and steganalysis can be seen as the same thing. Their purpose is to differentiate between tampered images and natural images. Modifications made to the specific attributes of an image will affect the corresponding residuals, as the residuals are closely linked to those attributes. SRM features are obtained by first capturing the essential noise characteristics with 30 basic filters. Then, nonlinear processes, such as selecting the maximum and lowest values from the neighboring outputs after filtering, are applied. The filters produce a quantified output that is then shortened by SRM. SRM then extracts the nearest co-occurrence data as the final features. This method produces a characteristic that can be regarded as a description of local noise.

### 3.2 Feature Enhancement Module

An improvement module is utilized to enhance the feature representation capability of ConvNext block and is applied to the feature maps. This deployment aims to utilize the high-frequency features that are retrieved from the appropriate layers. The features obtained from CNN layers are often noisy, leading to a potential degradation in performance. The deep layers exhibit a reduced amount of high-frequency information in their characteristics [24].

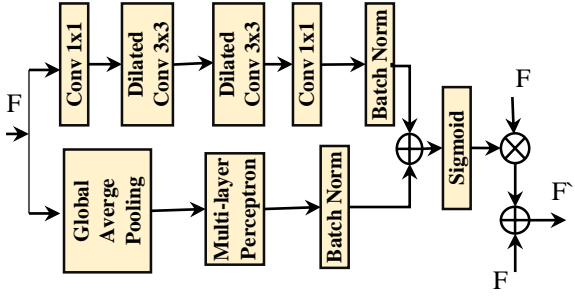

Figure III: Architecture of Feature Enhancement Module

The architecture of the feature enhancement module is shown in Figure III. With a feature map of $\mathcal{F} \in \mathbb{R}^{\mathcal{H} \times \mathcal{W} \times \mathcal{C}}$ outputted by the ConvNext block, the dilated convolution is used to enlarge the receptive field and $conv_{3 \times 3}^{dilated}$ and the $conv_{1 \times 1}$ are used to adjust the channel dimension. The attention map generated from the above branch is as follows:

$$M_1(F) = BN(conv_{1 \times 1}(conv_{3 \times 3}^{dilated}(conv_{1 \times 1}(F)))) \quad (2)$$

where batch normalization is indicated by BN. Moreover, the channel attention is formed by utilizing the inter-channel link. The feature map in each channel is aggregated by applying global average pooling (GAP) to the channel feature map is created which is followed by the MLP layer and batch norm layer.

$$M_2(F) = BN(MLP(GAP(F))) \quad (3)$$

Next, we merge the matrices $M_1(F)$ and $M_2(F)$ by adding their corresponding elements together and then apply a sigmoid function to generate the ultimate attention map.

$$M(F) = Sig(M_1(F) \oplus M_2(F)) \quad (4)$$

In the end, the features are improved by performing element-wise multiplication with the attention feature map and subsequently adding the result to the existing feature map.

$$F` = F \oplus F \otimes M(F) \quad (5)$$

### 3.3 Edge Extraction Block

Edge detection is a method employed to detect the areas in an image where there is a significant and sudden shift in brightness. The abrupt variation in the intensity value is detected at the points of lowest or highest values in the image histogram, by utilizing the first-order derivative. This change of gradient allows us to effectively locate the manipulation at edges. This consists of two operations or modules where first extraction information is fetched using the sobel operator which is further enhanced by the edge extraction module

*Sobel operator:* The edge can be determined by calculating the differentiation of pixel intensities. The Sobel mask calculates the first-order derivative and the edge is depicted by the local maximum or local minimum. The coefficients of the masks in the Sobel operator can be adjusted to meet our specific requirements, as long as they adhere to all the characteristics of derivative masks.

$$Sobel_{X-axis} = [-1\ 0\ 1; -2\ 0\ 2; -1\ 0\ 1] \quad (6)$$
$$Sobel_{y-axis} = [-1\ -2\ -1; 0\ 0\ 0; 1\ 2\ 1] \quad (7)$$

*Edge extraction block:* The task of detecting image manipulation involves identifying extremely faint indications of alteration inside the image. Furthermore, the subtle distinctions between the manipulation edge and the surrounding non-manipulation region are of great significance. To more effectively capture this nuanced information, it is necessary to preserve the features in the Convolutional Neural Network (CNN) at a rather high resolution.

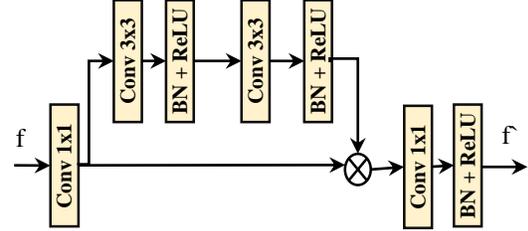

Figure IV: Architecture of Edge Extraction Block

As each layer in the Convolutional Neural Network (CNN) learns distinct feature contents, extract the manipulation edge by using the features produced by each layer of the network. To enhance the extraction of edges, we used the implementation of a specifically engineered edge extraction block (EEB). Figure IV illustrates the progression of EEB. To optimize computation and maximize the utilization of feature information, we employ a $1 \times 1$ convolution operation to decrease the number of channels in the features by a factor of 1/4. Subsequently, we proceed with constructing residual learning. Ultimately, a $1 \times 1$ convolution is employed to decrease the number of channels to 1.

$$M(f) = conv_{1 \times 1}(conv_{3 \times 3}(BN(ReLU(conv_{3 \times 3}(BN(ReLU(f))))))) \quad (8)$$

The above attention gets multiplied to further improve the edge feature representation. These features contain a rich presentation of the edge features.

$$f = conv_1 \otimes M(f) \quad (9)$$

### 3.4 ConvNext Module

The paper's authors present ConvNext, a new convolutional network design that replaces conventional "Conv2d" layers with modern ones influenced by recent breakthroughs in Vision Transformers (ViTs)[1]. The authors suggest several novel design principles to enhance the basic Convolutional Networks. These include employing larger kernel sizes, utilizing GELU activation instead of ReLU, augmenting the depth-wise convolutions, and substituting batch normalization with layer normalization. Using larger kernel sizes allows for a better understanding of a wider spatial context, which leads to more accurate detection of subtle artifacts. Utilizing deeper convolutions enhances the knowledge of spatial context without significantly increasing computational cost. Additionally, incorporating layer normalization promotes more stable training, which is essential for detecting fine manipulations. The enhanced hierarchical architecture and activation function enable the acquisition of intricate features, encompassing both high-level contextual features and low-level intricate details that are highly responsive to subtle variations. Incorporating these design elements, the ConvNext architecture enables enhanced robust performance and highly comprehensive feature extraction capabilities that are essential to manipulation detection tasks.

### 3.5 Loss function

This section will provide an analysis of the loss function employed in our localization model. Three renowned loss functions for classification and segmentation tasks are Edge Loss(EL), Binary Cross Entropy (BCE) Loss, and Focal Loss (FL) used in our model.

*Edge Loss:* To address the observation that artifacts are generally more common near the periphery of tampered regions i.e. edges, where the distinctions between manipulated and real areas are most apparent, we have devised an approach that prioritizes the boundary region of the manipulated area. As non-edge pixels dominate the pixels of an edge, we employ the Dice loss[2] for manipulating edge detection, referred to as $loss_{edge}$. The model's edge information at various semantic levels is continuously concatenated, and the loss is computed by manipulating both the edge information and the manipulation mask information. By integrating edge information, the model can prioritize the segmentation border of the object and enhance the overall accuracy of localization.

*The BCE loss*, derived from the Bernoulli distribution, aims to quantify the disparity between the probability distributions of the predicted and actual masks [25]. Segmentation is commonly employed for classification purposes, as it involves classifying pixels at a granular level. It is characterized or described as:

$$\mathcal{L}_{BCE}(G_m, \hat{P}_m) = -(G_m \log(\hat{P}_m) + (1 - G_m) \log(1 - \hat{P}_m)) \quad (10)$$

Where $G_m$ refers to the ground truth pixel value of the mask and $\hat{P}_m$ is the predicted pixel value of the mask. The method calculates the loss for each pixel and assigns equal importance to every pixel.

*Focal Loss* is a commonly observed phenomenon in segmentation tasks and is used as an addition to Binary Cross Entropy (BCE) in detection tasks that have a severe imbalance in class distribution [26].

The Focal Loss method effectively guides the network to focus on difficult samples without the need for any weight adjustment. This enables the model to acquire knowledge about more intricate scenarios while reducing the significance of less intricate ones. The focal Loss method employs a modulating factor, denoted as p, to reduce the importance of easy examples and instead focus the training on challenging negative examples.

$$\mathcal{FL}_{\gamma,\alpha} = -(\alpha(1-p_i)^\gamma y_i \log p_i + (1-\alpha)p_i^\gamma (1 - y_i) \log(1-p_i)) \quad (11)$$

Where $p_i$ represents the estimated probability distribution for the class label, which is equal to one. An equilibrium factor is denoted as α, and the rate at which simple data samples are down-weighted is defined by the parameter γ.

**The combined loss function** is a potent amalgamation of Binary Cross Entropy (BCE), Edge Loss and Focal Loss. The BCE loss function penalizes deviations from a normal distribution in the data samples. On the other hand, the Focal Loss functions specifically tackle the issue of class imbalance and prioritize the most challenging aspects of the task by reducing the impact of easier ones through their hyper parameter settings. The idea of edge loss specifically targets the border regions of manipulation, which plays a critical role in achieving accurate localization outcomes. This combination helps in overall accuracy precision while maintaining the boundary precision.

$$\mathcal{L}_{combined} = \mathcal{L}_{BCE} + \mathcal{FL}_{\gamma,\alpha} + loss_{edge} \quad (12)$$

Equation (12) denotes the comprehensive loss function employed in this model. The model utilizes this integrated loss function during the comprehensive training of the model.

### 4 Experiments

The purpose of the current section is to examine the effectiveness of the proposed methodology by validating it on a number of benchmark datasets and comparing it to a number of different cutting-edge manipulation localization techniques.

#### 4.1 Datasets
#### 4.1.1 Shallowfake dataset

The following datasets were utilized for the purpose of using the proposed model for training and validation:

- **CASIA** [27]. CASIA v1.0 comprises a total of 920 counterfeit images, which have been predominantly manipulated by splicing and copy-move techniques. CASIA v2.0 is an enhanced iteration of CASIA v1.0. However, the latter comprises 5063 counterfeit images, many of which feature more complex modifications suitable for network training. CASIA includes both CASIA v1.0 and CASIA v2.0. The CASIA v2.0 dataset was utilized for training, while the CASIA v1.0 dataset was employed for testing.

- **NIST16** [28]. NIST16 includes 564 images and encompasses three manipulation techniques: copy-move, splicing, and removal. NIST16 is a dataset that presents a challenge due to the post-processing that has been implemented to obscure any indications of potential image manipulation.
- **COLUMBIA** [29]**.** The Columbia dataset employs splicing alteration, which involves a total of 180 photos. Spliced images are created by directly copying and pasting visually hidden elements from Adobe Photoshop onto the original photographs, without any additional editing or modifications.
- **COVERAGE** [30]**.** A 100 images manipulated smaller dataset designed by copy-move operation and their corresponding Ground truth masks are included. To get rid of visible traces, every image is post-processed to hide visual traces.

Table I provides information about the division of the dataset into training and testing sets, along with the specific methods used for manipulation. Deep learning network training is characterized by a significant appetite for data. The current datasets commonly employed for training deep neural networks in image alteration detection lack an adequate number of images. In addition, modified images from a typical dataset may not provide enough training material as they have fewer imperfections. The model is initially trained using CASIAv2 and then further refined by fine-tuning with additional datasets. Subsequently, testing is performed on the datasets that were previously mentioned.

Table I: The benchmark datasets are partitioned into separate sets for training and testing purposes. into training and testing sets.

| Datasets | Training Set | Validation Set | Testing test | Total Samples |
|---|---|---|---|---|
| CASIA V2.0 | 5063 | - | - | 5063 |
| CASIA V1.0 | - | 600 | 320 | 920 |
| NIST16 | 414 | 75 | 75 | 564 |
| COLUMBIA | 130 | 25 | 25 | 180 |
| COVERAGE | 60 | 20 | 20 | 100 |

### 4.1.2 Deepfake Dataset

Currently, there is no deepfake image dataset that includes an accurate mask indicating the areas that have been modified. Zhang et al. [31] have constructed their dataset using Faceforensics++ [32], At now, there is no existing deepfake image dataset that includes an accurate mask indicating the areas that have been modified. Zhang et al. [40] have constructed their dataset using Faceforensics++ [41], the sole deepfake dataset that includes masks for the majority of its films. The Famous Faceforensics++ dataset comprises 1000 authentic videos and 5000 altered videos created using various techniques such as Deepfakes, Face2Face, Face-Swap, and Neural-Textures. Four operations are being evaluated for the frames recovered from the 1000 face shifter recordings, as these videos do not have any ground truth mask. Two frames are retrieved from each video, but certain accessibility problems have prevented us from downloading certain real and fake videos. We have extracted a total of 8,449 real frames and 7,330 fake frames[3].

### 4.2 Experimental setup

To construct our model, we have utilized the PyTorch framework. The model is executed on a pair of NVIDIA RTX A5000 graphics processing units (GPUs), and the image is resized to dimensions of 256 × 256. The model is optimized using the Adam optimizer with a batch size of 16 during both the training and testing phases. The initial learning rate is set to 1e-4 and it decreases by a factor of 0.8 every 10 steps. The model undergoes pre-training for 150 epochs and subsequently undergoes fine-tuning for an additional 50 epochs.

### 4.3 Evaluation Metric

The F1 score at the pixel level and the area under the receiver operating characteristic curve (AUC) are used as assessment metrics for comparison methods to quantify the localization performance. The more significant value indicates improved performance. The range of both the F1 score and the pixel-level AUC is [0,1]. These two metrics are commonly used for evaluation and comparison. For ablation studies and deepfake dataset comparison, we have used another metric named IoU (Intersection over Union).

### 4.4 Quantitative Analysis

In this section, we have conducted a quantitative analysis of the performance of the shallow fake dataset as well as of the deepfake dataset.

### 4.4.1 Shallowfake dataset

Based on the method described in reference [33], the model was trained using the CASIA2 dataset and further refined utilizing well-known shallowfakes datasets such as Nist16, Coverage, Columbia, and CASIA1. The Area Under the Curve (AUC) and F1 scores of these datasets have been documented in Table II. To provide a comparison, we will investigate two groups of models: unsupervised models and DNN models. The data presented in the table demonstrates that the model surpasses the unsupervised models by a substantial margin. The effectiveness of manually designed features tailored for a particular type of manipulation is greatly limited, and all of these conventional techniques only extract certain signs of tampering with little data for detection. The performance of our method is superior to previous DNN-based methods across several datasets, making them directly comparable. Our model demonstrates exceptional performance on the CASIA and Columbia datasets, with an impressive AUC score of 97.25%. Additionally, on the Nist16 dataset, our model's score is comparable to other models, with an AUC score of 99.79%. The incorporation of edge supervision of the model allows for handling complex cases of thin structures or small objects where edge information can be vital and leads to the overall improvement of results.

By contrast, the model exhibits subpar performance on the COVERAGE dataset, achieving an AUC score of

88.26% due to the presence of fewer images and the inclusion of duplicated or relocated objects that have a similar appearance. Furthermore, the dataset encompasses a distinct range of manipulation scenarios, including variations in lighting conditions, background perspectives, and object appearances. These factors pose a challenge for our model to effectively generalize, particularly when trained on a less diverse dataset. Our technique efficiently gathers a diverse set of data, encompassing RGB attributes, noise inconsistencies, and global context, instead of just depending on adjacent pixels. This enables us to obtain more extensive data for categorization and analysis. The ConvNext module enables the modelling of features at different semantic scales, allowing it to capture both low-level manipulations such as blobs and high-level semantic aspects such as texture. Therefore, the inclusion of multi-scale characteristics allows the model to better prioritize pixel-level image segmentation. The poor performance of complex CNN models can be attributed to the use of DNN-based techniques that use several CNN networks or complex branches to model the network. TDA-Net [33] is an exemplification of a model that combines three distinct CNN streams. This integration entails the training of a complex network in a holistic manner, which presents problems in terms of training complexity and heightened computational demands. Furthermore, there exists only a restricted range of models that are significantly less intricate, as they exclusively prioritize semantic information, leading to an incapacity to accurately detect manipulated sections. However, our model is less complex in comparison, skilled at capturing non-semantic features, and does not require a large quantity of training data to obtain equal performance. Therefore, this technique allows the model to effectively prioritize pixel-level image segmentation by capturing multi-scale information.

Table II: Results of shallowfake dataset model evaluation tests. "-" indicate an unknown score.

| Category | Method | COVERAGE | | NIST16 | | CASIA | | COLUMBIA | |
|---|---|---|---|---|---|---|---|---|---|
| | | AUC | AUC | AUC | AUC | AUC | F1 | AUC | F1 |
| Unsupervised | ELA [34] | 0.583 | 0.583 | 0.429 | 0.613 | 0.613 | 0.214 | 0.581 | 0.222 |
| | NOI1 [35] | 0.587 | 0.587 | 0.487 | 0.612 | 0.612 | 0.263 | 0.546 | 0.269 |
| | CFA1 [36] | 0.485 | 0.485 | 0.501 | 0.522 | 0.522 | 0.207 | 0.720 | 0.190 |
| DNN based methods | LSTM-EnDec [10] | 0.712 | 0.712 | 0.794 | - | - | - | | - |
| | Transforensics [21] | 0.884 | 0.884 | - | 0.850 | 0.850 | 0.627 | - | 0.674 |
| | ManTra-Net [11] | 0.819 | 0.819 | 0.795 | 0.817 | 0.817 | - | 0.824 | - |
| | TDA-Net [33] | 0.864 | 0.864 | 0.948 | 0.831 | 0.831 | 0.582 | 0.892 | 0.474 |
| | MFCN [9] | - | - | - | - | - | 0.541 | - | - |
| | SPAN [41] | 0.937 | 0.937 | 0.961 | 0.838 | 0.838 | 0.382 | 0.936 | 0.558 |
| | GSR-Net [40] | 0.768 | 0.768 | 0.945 | 0.796 | 0.796 | 0.574 | | 0.489 |
| | ObjectFormer [20] | 0.957 | 0.957 | 0.996 | 0.882 | 0.882 | 0.579 | - | 0.7580 |
| | CR-CNN [39] | 0.939 | 0.939 | 0.992 | 0.789 | 0.789 | 0.475 | 0.861 | 0.757 |
| | J-LSTM [38] | 0.712 | 0.712 | 0.764 | - | - | - | - | - |
| | PSCC-Net [42] | 0.941 | 0.941 | 0.996 | 0.875 | 0.875 | 0.554 | - | 0.723 |
| | MVSS-Net++ [19] | 0.897 | 0.897 | 0.976 | 0.844 | 0.844 | 0.546 | - | 0.753 |
| | RGB-N [37] | 0.817 | 0.817 | 0.937 | 0.795 | 0.795 | 0.582 | 0.858 | 0.474 |
| | TA-Net [43] | **0.978** | **0.978** | 0.997 | 0.893 | 0.893 | 0.614 | - | **0.782** |
| | **Our Model** | 0.8826 | 0.8826 | **0.9979** | **0.9542** | **0.9542** | **0.8956** | **0.9725** | 0.6075 |

#### 4.4.2 Deepfake dataset

Ten models are being assessed for their performance on the deepfake dataset. There are six state-of-the-art image manipulation models, while the remaining models are typical image-segmentation models. The codes of the models available on GitHub are taken into consideration for comparison. PyTorch models that have been pre-trained are utilized for image segmentation models, and these models are then fine-tuned. Three assessment measures are employed to provide a more thorough and all-encompassing examination. Table III displays the empirical findings of the deepfake technique on several models. All the models exhibited satisfactory performance, except MantraNet [11], across the several categories of the Faceforensics++ dataset. This phenomenon can be ascribed to the prevalence of facial modifications, the presence of a sole entity occupying the complete frame, and the capability of all the models to effectively represent these noticeable imperfections. MantraNet exhibits subpar performance across all deepfake categories, likely due to the low resolution, blurriness, and noise present in the images/frames. This could be attributed to the model's heavy reliance on the diverse and enough training dataset to perform well and the absence of which degrades the model performance badly. NedB-Net [44] is a model that has achieved good performance, although its score is considerably lower compared to other state-of-the-art (SoTA) models. This phenomenon can be ascribed to the existence of substandard images and more conspicuous modified areas, as well as the model's susceptibility to different types of noise and edge patterns. The authors of the research acknowledge this issue. DL-Net [45] The performance of DL-Net on the FF++ dataset is quite good, achieving an F1 score of 96%. This is due to its ability to effectively capture both high and low-level cues by predicting noise-level segmentation maps. These maps help the model to focus on the regions that have been altered specifically. Nevertheless, the model's performance in the category of Face-swap manipulation could be better compared to the other three categories of FF++ manipulation. This is because face-swap techniques sometimes involve applying smoothing or blurring effects, which alter the noise and semantic patterns that the model depends on to detect manipulation. Another technique [46] employed for deepfake localization is a weak supervision framework that utilizes three methods: GradCAM, Patches, and Attention, to illustrate the results. We employed GradCAM techniques to compare scores. The technique excels in the weak supervision environment, demonstrating the model's strong discriminative skills. However, similar to the previous version, the model experiences a decrease in performance when it comes to the FS category of manipulation. This could be attributed to the fact that the technique is specifically designed for diffusion-generated images, which may restrict its capacity to perform effectively on GAN-generated images. Additionally, another technique [31] showed excellent performance, with an F1 score of 98%. Their approach is based on the preexisting UperNet and uses Bayar convolution techniques to detect and track noise indicators. Although these models are considered state-of-the-art, all of them achieve a score higher than 90%. This can be ascribed to the fact that much of the alteration has been focused on the face, which is easily recognizable by the models. The DADF technique [47] outperforms most models by utilizing multi-scale adapters to detect both short and long-range forgeries, as well as guided attention mechanisms that enhance the identification of rich forgery clues. Their scores are comparable to other approaches and serve as a suitable benchmark for comparison with a state-of-the-art method. Our model has demonstrated significant performance in the different categories of the FF++ dataset and surpassed the scores of other conventional models. The evaluations demonstrated that the effective coordination of multiple modules led to the development of a strong ability to discern and adapt to various forms of manipulation.

Table III: Test results of several models on FF++ dataset categories. IMD and IS are Image Manipulation Detection models, and Image Segmentation Models

| Types | Methods | DeepFakes | | | Face2Face | | | FaceSwap | | | Neural Texture | | |
|---|---|---|---|---|---|---|---|---|---|---|---|---|---|
| | | IoU | AUC | F1 | IoU | AUC | F1 | IoU | AUC | F1 | IoU | AUC | F1 |
| IMD Models | DADF [47] | 0.9453 | 0.9939 | 0.9677 | 0.9621 | 0.9899 | 0.9786 | 0.9599 | 0.9896 | 0.9655 | 0.9236 | 0.9788 | 0.9586 |
| | DL-Net [45] | 0.8750 | 0.9952 | 0.9337 | 0.9108 | 0.9946 | 0.9533 | 0.8976 | 0.9976 | 0.9460 | 0.9262 | 0.9979 | 0.9617 |
| | MVSS-Net [48] | 0.9558 | 0.9996 | 0.9787 | 0.9788 | 0.9997 | 0.9893 | 0.9570 | 0.9989 | 0.9780 | 0.9382 | 0.998 | 0.968 |
| | NedB_Net [44] | 0.8811 | 0.9767 | 0.9368 | 0.8432 | 0.9662 | 0.9149 | 0.8522 | 0.9700 | 0.9221 | 0.8827 | 0.9783 | 0.9377 |
| | Weakly_Super_Gradcam [46] | 0.9787 | 0.9990 | 0.9897 | 0.8753 | 0.9795 | 0.9336 | 0.9575 | 0.9791 | 0.9231 | 0.9861 | 0.9990 | 0.9868 |
| | MantraNet [11] | 0.3469 | 0.9523 | 0.5151 | 0.3553 | 0.9159 | 0.5243 | 0.3394 | 0.8888 | 0.5068 | 0.3664 | 0.9706 | 0.5363 |
| | ShallowDeepfake_local [31] | 0.9617 | 0.999 | 0.9713 | 0.9801 | 0.9898 | 0.9799 | 0.9485 | 0.9856 | 0.9666 | 0.9365 | 0.9989 | 0.9689 |
| IS Models | LRASPP [51] | 0.9114 | **0.9992** | 0.9536 | 0.9445 | 0.9998 | 0.9714 | 0.9106 | **0.9995** | 0.9532 | 0.9399 | 0.9997 | 0.9690 |
| | FCN [50] | 0.9701 | 0.9967 | 0.9848 | 0.9834 | 0.9988 | 0.9786 | 0.9470 | 0.9991 | 0.9728 | 0.9591 | 0.9984 | 0.9728 |
| | DeepLab [49] | 0.9428 | 0.9981 | 0.9706 | 0.9840 | **0.9999** | 0.9919 | 0.9769 | 0.9986 | 0.9883 | 0.9704 | **0.9998** | 0.9640 |
| | Our Model | **0.9812** | 0.9956 | **0.9907** | **0.9898** | 0.9989 | **0.9945** | **0.9896** | 0.9987 | **0.9943** | **0.9865** | 0.9978 | **0.9972** |

### 4.5 Qualitative Analysis

This section provides a comparison between our method and the two most competitive methods, namely MantraNet and MVSSS, to present detailed qualitative results on both shallowfakes and deepfake datasets. Figure V displays the visual representation of the outcomes obtained from the detection of image alteration. Our method outperforms other methods in terms of localization accuracy, as the other methods produce a significant number of false positives. Regarding the shallowfake dataset, the approach exhibits exceptional localization performance for the CASIA, Columbia, and Nist16 datasets. Localization for the deepfake dataset appears to be quite straightforward, given that most image alteration is focused on the face, simplifying the localization process. MantraNet deviates significantly from the ground truth, while MVSSNET shows a notable occurrence of false positives in unaltered regions. The main factor contributing to this result is that during the training phase, MVSSNET was exposed to a significant number of natural photos, which likely had a detrimental effect on the network's training process. Furthermore, the model requires a substantial amount of training data samples that exhibit a specific manipulation. Without this, the model's performance will decline. However, Mantra-Net faces difficulties in executing delicate modifications, such as methods that avoid producing abnormal artifacts or generating low-resolution images. These obstacles result in an increased number of false negatives for this model. Our approach prioritizes the analysis of low-level data such as noise, as well as high-level contextual variables with edge-based supervision, resulting in improved identification and localization of altered artifacts.

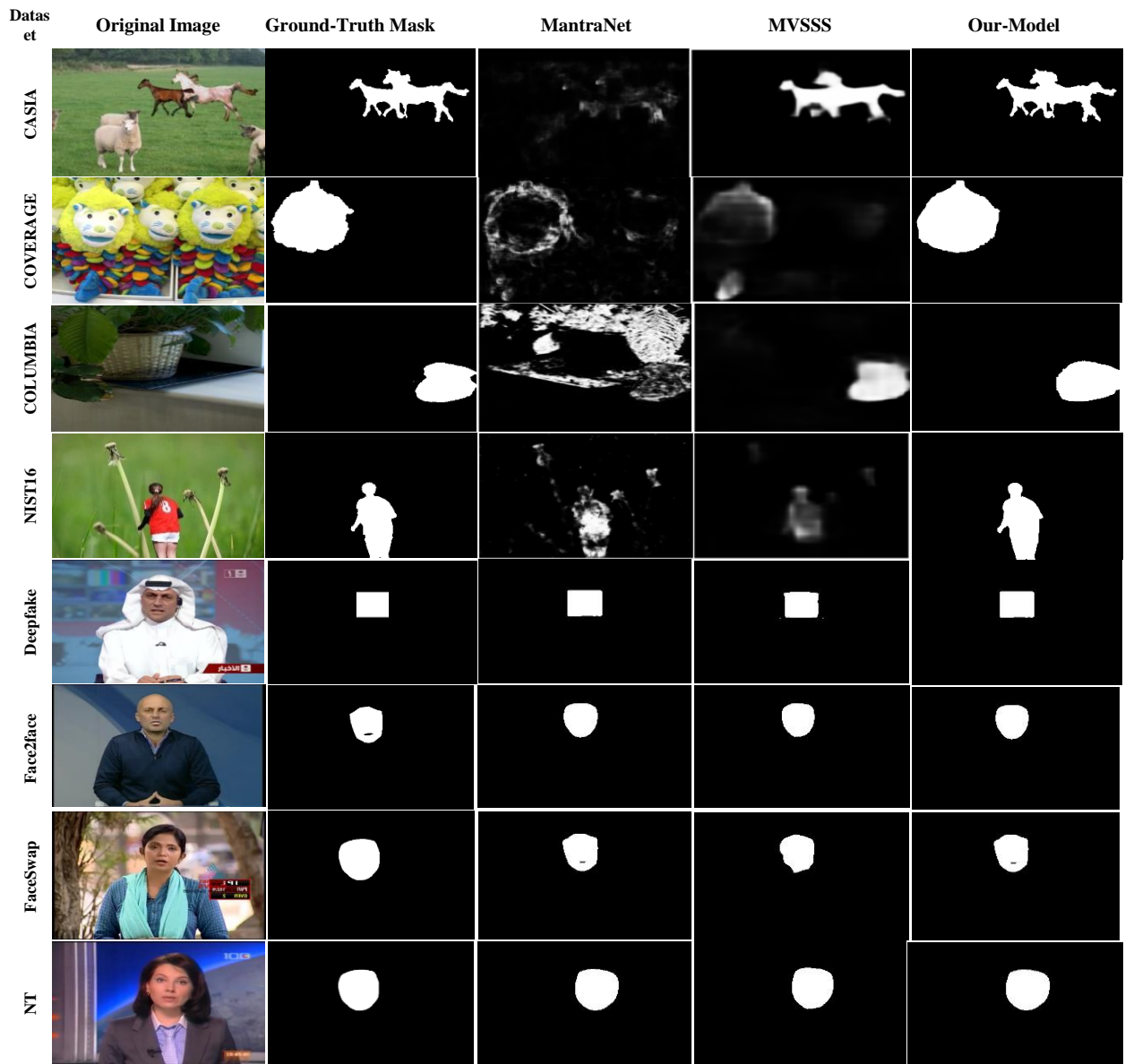

Figure V: Qualitative visualization of shallow fake and deepfake dataset images using different manipulation localization approaches.

### 4.6 Ablation Studies

We evaluate the proposed network in different scenarios by gradually introducing its components to analyze the effect of each component. The components underwent training using the CASIA2 dataset and were subsequently assessed using other shallowfake datasets, namely NIST16, Columbia, Coverage, and CASIA1. Table IV displays the outcomes of the ablations experiment. Below, we will discuss the various experimental configurations.

**Case A: Model without edge loss supervision:** Here the model is trained without edge supervision loss. It can be seen from the first row of the table that the model experiences a substantial decrease of 5-7% in the overall score of the evaluation metric, this suggests that edge supervision plays a vital in the overall detection accuracy of the model. Edge supervision allows the model to focus on the boundaries of the manipulation leading to precise manipulation results on the edges particularly, the manipulation involves finer details or thin structures. Also, edge supervision allows the model to learn to discriminate between true edges and noisy artifacts, which helps to recognize intricate structures.

**Model without feature enhancement module:** The second row represents the score and it can be seen there is a decrease of 3-4% in the overall score of localization. The phenomena can be attributed to the directing of attention of the model towards high-level or semantic features, which are essential for manipulation. In addition, the module consistently shows its effectiveness on many datasets, suggesting its ability to acquire a wide range of contextual information for varied manipulations. Moreover, it is noteworthy that the scores exhibited a slight and uniform decline across all the datasets.

**Model without Noise branch:** In this instance, the noise branch, is eliminated from the model. The detection mostly relies on the utilization of RGB high-level or semantic properties. In this instance, the performance witnessed a substantial decrease of around 6-8%, validating the important influence of low-level factors like noise consistencies on the overall detection task. The CASIA model has exhibited a substantial decline in scores, specifically around 15%, when exposed to various forms of image manipulation. This exemplifies the significance of the residual noise characteristic in identifying different types of manipulation.

**Model without RGB branch** The model's RGB component has been removed and the noise component is used for localization purposes. In such conditions, the AUC score undergoes a rather moderate decrease of around 3-5%. This further affirms that the low-level noise components are more important than high-level semantic information for the detection task. The performance of the Coverage dataset has experienced a substantial decrease, suggesting that a sufficient number of datasets for branches with noise is necessary to train and detect inconsistencies caused by noise efficiently.

Table IV Different component ablation experiments using CASIA2-trained and another dataset-tested model.

| Method | NIST16 | | COLUMBIA | | COVERAGE | | CASIA | |
|---|---|---|---|---|---|---|---|---|
| | AUC | F1 | AUC | F1 | AUC | F1 | AUC | F1 |
| Case A | 0.9085 | 0.9316 | 0.9016 | 0.8589 | 0.7645 | 0.6012 | 0.8916 | 0.8474 |
| Case B | 0.9316 | 0.9715 | 0.9519 | 0.8879 | 0.8052 | 0.6326 | 0.9278 | 0.8715 |
| Case C | 0.8706 | 0.9289 | 0.8996 | 0.8312 | 0.7889 | 0.6019 | 0.8195 | 0.7746 |
| Case D | 0.9236 | 0.9456 | 0.9356 | 0.8829 | 0.7815 | 0.6159 | 0.9056 | 0.8579 |
| Overall Model | 0.9512 | 0.9976 | 0.9746 | 0.9188 | 0.8366 | 0.6512 | 0.9524 | 0.8945 |

## 5 Conclusion

This study introduces a new dual-branch design that incorporates Noise Residual extraction modules in one branch and RGB information in the other branch. The architecture utilizes a conventional ConvNext module, whose features are further improved by the feature improvement module. The model additionally utilizes edge supervision to improve the localization of the manipulation specifically at the edges. The model accurately captures the fundamental inconsistencies that are crucial for interpretability in machine learning tasks, as well as incorporating further semantic variables. Extensive experimentation conducted on both shallowfakes and deepfake datasets has demonstrated that the model successfully detects minute indications of manipulation and delivers state-of-the-art (SoTA) outcomes. Future work may entail assessing the model's capacity to apply to new, unexplored data and doing a thorough review of different compression scenarios to ensure reliability.

Declaration

## 6 Ethical Approval
NA (Not Applicable)

## 7 Funding
No funding was received for this research.

## 8 Availability of data and materials
Data is open-source, and no prior permission is required. It is available through these links.
CASIA: casia dataset (kaggle.com)
Nist16: OpenMFC (nist.gov)
Coverage: wenbihan/coverage: Copy-Move forgery database with similar but Genuine objects. ICIP2016 paper (github.com)
Columbia: Columbia Image Splicing Detection Evaluation Dataset
Faceforensics++: ondyari/FaceForensics: Github of the FaceForensics dataset